\documentclass{article} 
\usepackage{iclr2026_conference,times}

\usepackage{amsmath,amsfonts,bm}

\def\eqref#1{equation~\ref{#1}}

\def\1{\bm{1}}

\DeclareMathAlphabet{\mathsfit}{\encodingdefault}{\sfdefault}{m}{sl}
\SetMathAlphabet{\mathsfit}{bold}{\encodingdefault}{\sfdefault}{bx}{n}

\usepackage{hyperref}
\usepackage{url}

\usepackage{graphicx}
\usepackage{stfloats}
\usepackage{listings}
\usepackage{xcolor}
\usepackage{float}

\usepackage{booktabs, tabularx, array, longtable}

\title{OMEGA: Optimizing Machine Learning by Evaluating Generated Algorithms}

\author{Jeremy Nixon \\
Infinity Artificial Intelligence Institute\\
San Francisco, CA, USA \\
\texttt{jeremy@infinity.inc} \\
\And
Annika Singh \\
Infinity Artificial Intelligence Institute \& \\
Computer Science, Stanford University \\
Palo Alto, CA, USA \\
\texttt{annikaks@{stanford.edu}} \\
\texttt{annikas@{infinity.inc}}\\
}

\iclrfinalcopy 
\begin{document}

\maketitle

\begin{abstract}
In order to automate AI research we introduce a full, end-to-end framework, OMEGA: Optimizing Machine learning by Evaluating Generated Algorithms, that starts at idea generation and ends with executable code. Our system combines structured meta-prompt engineering with executable code generation to create new ML classifiers. The OMEGA framework has been utilized to generate several novel algorithms that outperform scikit-learn baselines across a robust selection of 20 benchmark datasets (\texttt{infinity-bench}). You can access models discussed in this paper and more in the python package: \texttt{pip install omega-models}. 
\end{abstract}

\section{Introduction}
The evolution of machine learning (ML) models has historically been driven by the manual derivation and implementation of novel algorithms. However, the transition from a theoretical hypothesis to a production-ready, validated implementation remains a high-friction process. Translating a novel, non-intuitive idea into executable code often requires extensive debugging for both coding and integration within existing pipelines. While existing automation techniques, such as Neural Architecture Search (NAS) and AutoML, have succeeded in optimizing hyperparemeters and layer choice within fixed layer types, the discovery of entirely new algorithmic logic remains largely a manual endeavor \citep{Zoph2016Neural, Real2020AlphaEvolve}.

In this paper, we propose a shift in how we utilize Large Language Models (LLMs) for machine learning research. Instead of treating LLM outputs as static text artifacts, we treat them as executable learning systems. We investigate whether LLMs can reason about and generate novel algorithms that are competitive with established baselines without human intervention. To facilitate this, we introduce \textbf{OMEGA}, a framework that specializes LLM synthesis for the creation of standardized, API-compliant machine learning estimators. OMEGA bridges the gap between raw code generation and systematic algorithmic evaluation, enabling a closed-loop system for algorithmic discovery.

\begin{figure*}[hb] 
    \centering
    \includegraphics[width=1\linewidth]{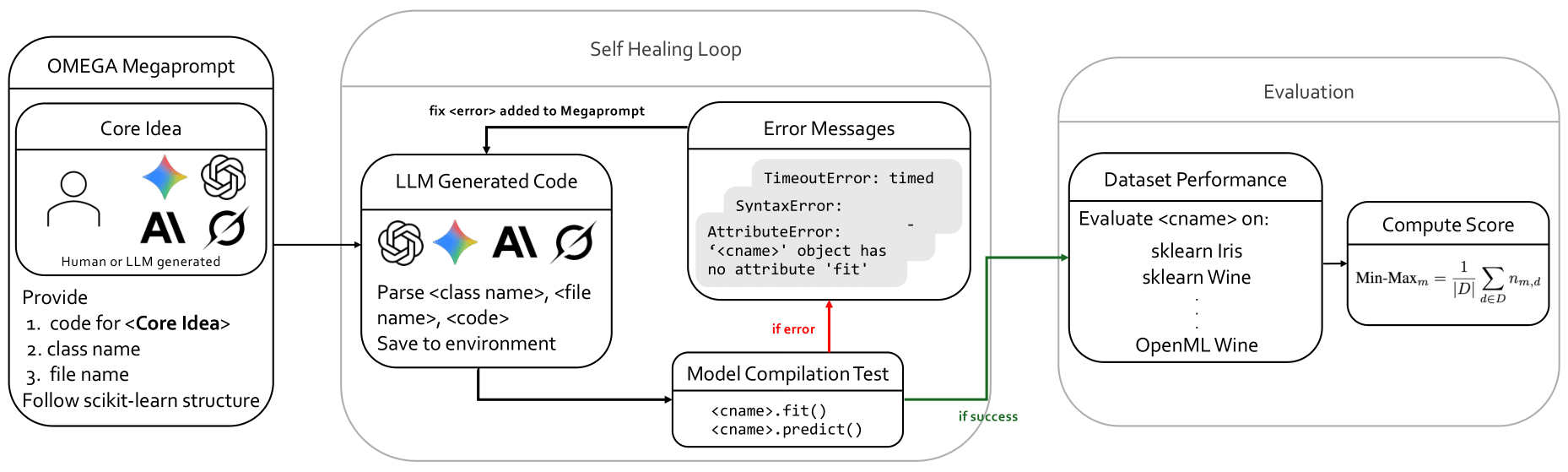}
    \caption{Core OMEGA Framework}
    \label{fig:omega_main_img}
\end{figure*}

\subsection{Detailed Contributions}
Our contributions are fivefold 
\begin{enumerate}
    \item We propose OMEGA: an automated end to end framework that allows developers to enter a simple prompt for a novel classification model and return compile-error free, scikit-learn compatible code that has been evaluated on our benchmarking dataset. 
    \item We propose \texttt{infinity-bench}: A benchmark to evaluate classifcation models on robustness and accuracy. 
    \item We analyze 2 (of many) novel classification models generated using OMEGA, that out perform scikit-learn baselines.
    \item We present an analysis comparing 4 popular LLMs' coding ability. 
    \item We present results regarding recursive self prompting and code improvement across 4 LLM's used in OMEGA.
\end{enumerate}

\subsection{Brief Literature Review}
OMEGA is at the intersection of automated discovery, meta-learning, and neural program synthesis.

\textbf{AutoML and Architectural Discovery:} Traditional AutoML has primarily focused on structured search over fixed sets of algorithms and hyperparameters to optimize performance for specific data manifolds \citep{Hutter2019Automated}. Early efforts in algorithmic generation utilized genetic programming and evolutionary algorithms to evolve programs via natural selection, though these methods often struggled with the complexity of modern ML architectures \citep{Koza1992Genetic}. This led to the rise of Meta-learning and Neural Architecture Search (NAS), where systems optimize learning strategies using reinforcement learning and Bayesian optimization \citep{Zoph2016Neural, Snoek2012Practical, Finn2017Model, elsken2019neuralarchserach}.

\textbf{Autonomous Algorithmic Discovery:} Recent work has pushed discovery beyond architectural tuning toward the creation of new mathematical logic. AlphaEvolve demonstrated that machine learning algorithms could be evolved from basic primitives \citep{Real2020AlphaEvolve}. In parallel, systems like AlphaTensor and FunSearch have discovered non-intuitive and provably correct algorithms for fundamental tasks by combining deep learning with automated evaluators \citep{Fawzi2022AlphaTensor, Romera2024FunSearch}. Furthermore, OMEGA is inspired by the emergence of ``AI Scientists", which have suggested a future where the entire research pipeline, from hypothesis generation to paper writing, is fully automated \citep{Lu2024AIScientist, Akiba2024Sakana}.

\textbf{Program Synthesis via LLMs:} Current advances in neural program synthesis have transitioned from simple code completion to functional, natural-language-driven logic generation. Benchmarks such as HumanEval have established the baseline for functional correctness in LLM-generated code \citep{chen2021evaluating}. Modern execution environments now enable iterative debugging and closed-loop execution within the model's workflow \citep{openai2023code, openai2023assistants}. OMEGA builds on this shift by directing this generative capacity toward the synthesis of industry-standard, scikit-learn-compatible machine learning models.

\section{Methods}
The OMEGA framework mirrors a code generation framework many engineers manually utilize. We ingest a prompt that is either entered by a user or LLM generated using some initial model inspiration. This is used to generate code that will go through a self-healing pipeline to ensure execution capability before being evaluated on our benchmark. We dive into each of these steps below. 

\subsection{Idea Generation}
To generate executable algorithms, we adopt two complementary approaches: autonomous hypothesis generation by language models and human-submitted algorithmic ideas. \\

\textbf{LLM Ontology Search: }To trigger an LLM generating ideas itself, we provided a list of Models and Research Principles. For each of the models provided, we prompted the LLM to use the principles to generate 10 unique, novel ways to modify the base models. We parse through the list of ideas and use them as individual prompts for actual code generation. This mirrors prior work on meta-learning and computational creativity, where systems explore structured spaces of solutions \citep{Vilalta2002Perspective, Colton2008Creativity}.

\textbf{Human Prompted Ideas: }Simultaneously, we allowed humans to submit prompts to OMEGA regarding novel ideas they had for new models.
These ideas were entered in the framework at the same point as the parsed ideas from LLM generation.

\subsection{Code Generation}

Once the LLMs returned the responses for the prompts, the responses were parsed to extract the executable code. During this stage, we prioritized two things (1) easy integration with existing pipelines and (2) eliminating the need to revise the algorithm. 

\textbf{Scikit-Learn Design Patters: }For the former goal, we enforce scikit-learn design patterns by requiring generated models to inherit from \texttt{BaseEstimator} and implement \texttt{.fit()} and \texttt{.predict()}. This aligns with established API conventions that prioritize composability, evaluation consistency, and reproducibility \citep{Pedregosa2011Scikit, Buitinck2013API}. We treat scikit-learn as a domain-specific language (DSL) for algorithm generation. Thus, when using the algorithms from the package, it is easy to integrate into existing scikit-learn workflows.

\label{self-healing}
\textbf{Self-Healing: }For the latter goal, the framework includes a self-healing mechanism in which the error stacktrace is captured and fed back into the generation loop. This is inspired by similar execution-based validation strategies that have proven critical for reliable code synthesis and reproducible research \citep{Gundersen2018Reproducible, Aho2006Compilers, chen2021evaluating}. After a fixed set of retries, if the code fails to self heal, we don't return any code and instead ask the user to try again. This process is required to ensure that all classifiers that we publish are error-free, as unexamined code is often useless \citep{Wang_2025_unexamined_code}.

\subsection{Evaluation and \texttt{Infinity-Bench}}
Each valid model is evaluated on 20 classification datasets sourced from scikit-learn and OpenML \citep{Vanschoren2014OpenML}. To account for the variation in the datasets, we use min-max normalized accuracy and aggregate performance between datasets, following best practices for multi-data set evaluation \citep{Demvsar2006Statistical}.

\textbf{Performance Score: }Performance is purely evaluated on the basis of classification accuracy. However, we quickly found that certain included datasets were easier to perform well (many models were able to achieve near 100\% accuracy). Since our datasets have varying difficulty levels, we rank ranking algorithms relative to each other on each dataset, then aggregating these rankings \citep{brazdil2009metalearning}. This allows us to compare relative performance rather than absolute performance to account for dataset difficulty. The accuracy for each model on each is calculated as follows:

$$n_{m,d} = \frac{s_{m,d} - \min_d}{\max_d - \min_d} $$
Where $s_{m,d}$ is the accuracy of the model on dataset d, and $min_d$ and $max_d$ are the scores of the worst and the best models, respectively. We then take the average score of the model across all 20 datasets to compute the min-max score per model:
$$\text{Min-Max}_m = \frac{1}{|D|} \sum_{d \in D} n_{m,d}$$

\textbf{Dataset Diversity: }By only focusing on classification, we were able to ensure 
\begin{enumerate}
    \item models were properly formatted (important for the self healing property)
    \item models were easily comparable
\end{enumerate}
However, within classification, we ensured that we had dataset diversity via numerical and categorical features, varying dataset size, binary and multiclass data that spanned various fields including bio, medicine, education, etc. 

While our models haven't been tested on differing types of datasets (i.e., image/video), the core OMEGA framework can virtually be abstracted to any use case and is only constrained by the LLM's ability to handle complexity. While we recommend using the models presented in this paper and in our python package for classification tasks, we would encourage users to utilize the OMEGA framework for any other use cases. 

\textbf{Benchmark Dataset} 
With recent agentic developments, particularly with coding agents, algorithmic development within existing environments has become far easier. In order to establish a universal way to test classification model creation, we propose the \texttt{infinity-bench}, a repository containing our these datasets that can be used to test newly generated classification models on and evaluate and compare their performance. 

\subsection{Library Creation}
OMEGA democratizes the models that are created by including the top models in a python package (\texttt{omega-models}) that can be easily imported and utilized on other datasets. By ensuring that models take the format of scikit-learn packages, these imports can be seamlessly integrated with existing workflows.

\section{OMEGA-Generated Model Results}
\subsection{Top Generated Models}
\begin{figure*}[t] 
    \centering
    \setlength{\tabcolsep}{3pt}
    \begin{minipage}{0.48\textwidth} 
        \centering
        \includegraphics[width=\linewidth]{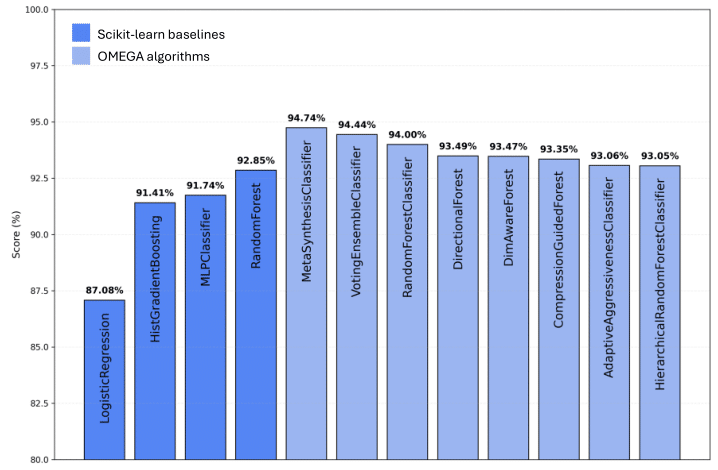}
        \label{fig:performance_barchart}
    \end{minipage}%
    \hfill%
    \begin{minipage}{0.48\textwidth} 
        \centering
        \begin{small}
        \begin{tabular}{llcc}
            \toprule 
            Rank & Model & MinMax & Generator \\ 
            \midrule 
            1 & MetaSynthesisClassifier & 0.9474 & User \\
            2 & VotingEnsembleClassifier & 0.9445 & User \\
            3 & RandomForestClassifier & 0.9401 & User \\
            4 & DirectionalForest & 0.9350 & System \\
            5 & DimAwareForest & 0.9347 & System \\
            6 & CompressionGuidedForest & 0.9336 & System \\
            7 & AdaptiveAggressivenessClassifier & 0.9307 & System \\
            8 & HierarchicalRandomForestClassifier & 0.9305 & User \\
            ... & ... & ... & ... \\
            11 & \textcolor{blue}{RandomForest} & \textcolor{blue}{0.9285} & \textcolor{blue}{Scikit-Learn} \\
            19 & \textcolor{blue}{MLPClassifier} & \textcolor{blue}{0.9174} & \textcolor{blue}{Scikit-Learn} \\
            21 & \textcolor{blue}{HistGradientBoosting} & \textcolor{blue}{0.9141}& \textcolor{blue}{Scikit-Learn} \\
            59 & \textcolor{blue}{LogisticRegression} & \textcolor{blue}{0.8708} & \textcolor{blue}{Scikit-Learn} \\
            \bottomrule 
        \end{tabular}
        \end{small}
    \end{minipage}
    \vspace{3mm}
    \caption{Best Models vs Scikit-Learn Baselines (Min-Max Score)}
    \label{fig:best models}
\end{figure*}

Above (Figure \ref{fig:best models}) we include the top performing algorithms as well as their aggregate score compared to the scikit-learn baseline algorithms (shown in blue). 

These figures demonstrate the viability of the OMEGA framework for the ideation and creation of classification modules. Below we do a deeper dive into two of the generated algorithms, one that is human prompted (\#1. $\texttt{MetaSynthesisClassifier}$ Section \ref{meta synthesis algorithm}) and one that was autonomously LLM prompted ($\#4.$ \texttt{Directional Forest} Section \ref{directional forest algorithm}) to better understand the novelty of algorithms generated by OMEGA). Models generated by Anthropic's Claude Sonnet 4.5.

\begin{figure*}[hb]
    \centering
    \includegraphics[width=0.9\linewidth]{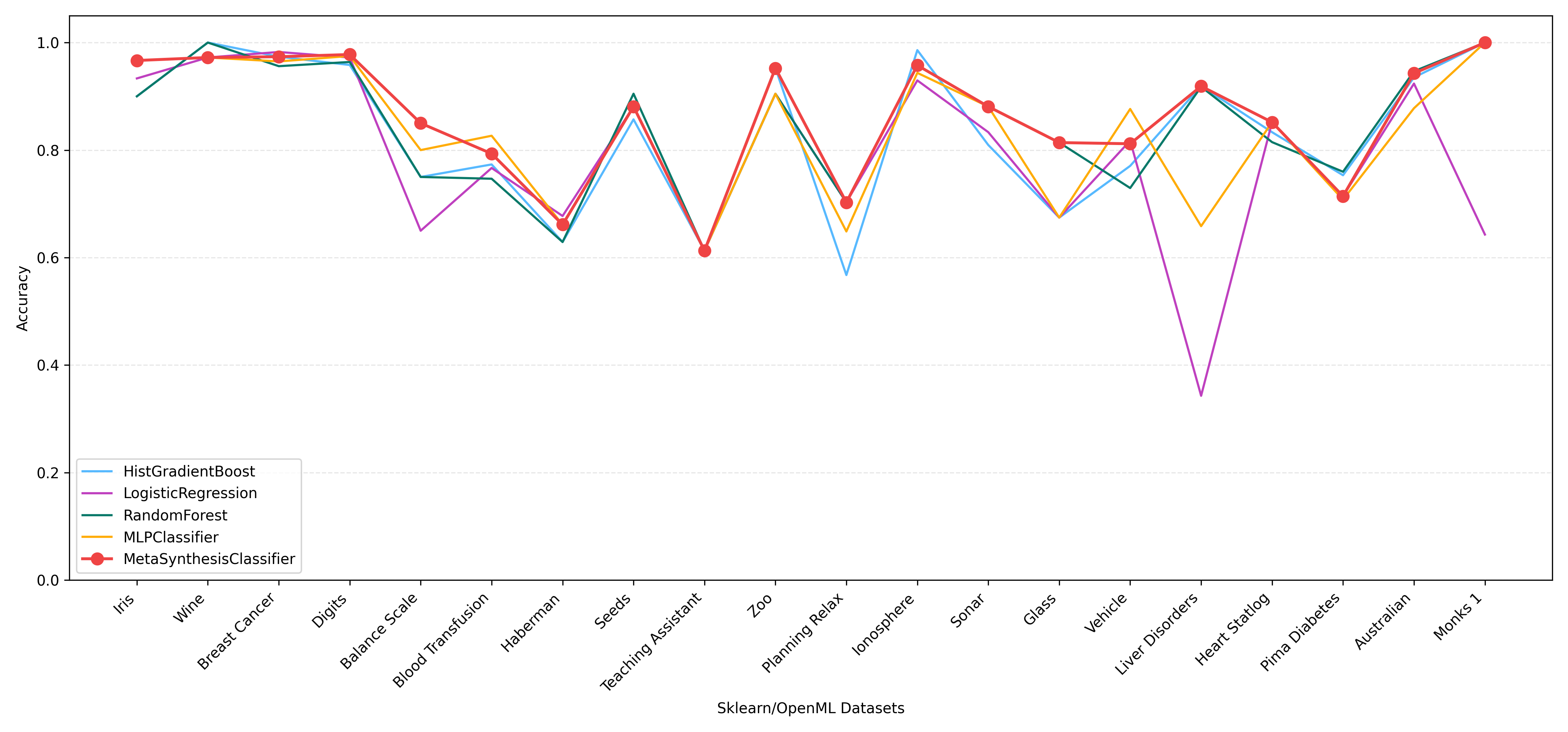}
    \caption{\texttt{MetaSynthesisClassifier} vs Scikit-Learn Individual Dataset Scores}
    \label{fig:meta_performance_trendline}
\end{figure*}

\subsection{\texttt{MetaSynthesisClassifier}}
\label{meta synthesis algorithm}
The \texttt{MetaSynthesisClassifier} represents a meta-learning approach to ensemble synthesis, specifically utilizing a stacked generalization architecture. Stacking is an ensemble technique in which a meta-learner is trained to optimally combine the predictions of several base models to improve generalization \citep{Wolpert1992Stacked}. The core objective is to have base learners and then treat the combination of these learners as a \textit{secondary} supervised learning task. The code can be found in Appendix \ref{app:metasynthesis_source_code}.

\subsubsection{Algorithm}
We define the system as a hierarchical set $\mathcal{H} = \{\mathcal{E}, M_{\psi}\}$, where $\mathcal{E} = \{E_1, E_2, \dots, E_m\}$ is a set of heterogeneous base estimators (e.g., Logistic Regression, Random Forest, and Decision Trees) and $M_{\psi}$ is the meta-estimator responsible for optimal synthesis. 

\textbf{Meta-Feature Generation:} For a training sample $(\mathbf{x}_i, y_i)$, the meta-feature vector $\mathbf{z}_i$ is constructed by concatenating the probability outputs of the base estimators, essentially storing the opinion of all the base estimations:
\begin{equation}
    \mathbf{z}_i = [P(y \mid \mathbf{x}_i; E_1^{(-k)}), \dots, P(y \mid \mathbf{x}_i; E_m^{(-k)})]
\end{equation}
where $E_j^{(-k)}$ is trained on a subset of data that excludes sample $i$. This is mainly done to prevent data leakage and ensure the meta-classifier will not inherit the training-set bias of base learners and \textit{actually} learn the relative reliability and error correlations of the generated algorithms based on their performance on previously unseen instances.

\textbf{Meta-Level Synthesis:} Once the meta-features $\mathbf{Z}$ are generated for the entire training set, the meta-estimator $M_{\psi}$ is trained on the mapping from the base learners' collective predictions to the true labels. This optimization problem is defined as:
\begin{equation}
    \min_{\psi} \sum_{i=1}^n \mathcal{L}(y_i, f(\mathbf{z}_i; \psi))
\end{equation}
This allows $M_{\psi}$ to learn which base learners are most reliable for specific patterns in the prediction space, effectively weighting their influence based on their historical accuracy during the cross-validation stage

\textbf{Final Inference: }During inference, a test sample $\mathbf{x}^*$ is passed through the fully-trained base ensemble $\mathcal{E}$ to produce the latent vector $\mathbf{z}^*$. The final prediction $\hat{y}$ is then derived from the meta-classifier's evaluation of that synthesized vector:
\begin{equation}
    \hat{y} = \text{argmax}_{c \in C} P(y=c \mid \mathbf{z}^*; M_{\psi})
\end{equation}
By learning to synthesize the diverse biases of the base learners, the \texttt{MetaSynthesisClassifier} achieves a lower generalization error than individual learners by finding the optimal manifold in the prediction space.

\subsubsection{Results}
As seen in Figure \ref{fig:meta_performance_trendline} \texttt{MetaSynthesisClassifier} demonstrates exceptional robustness. By adaptively learning to combine the strengths of linear models and tree-based ensembles, it successfully navigates different manifold complexities across the 20 benchmark datasets.

\subsection{\texttt{DirectionalForest}}
\label{directional forest algorithm}
The \texttt{DirectionalForest} incorporates feature directionality into an ensemble of decision trees to align the input space with class-specific statistical deviations. This algorithm was generated by the OMEGA framework to investigate whether pre-calculating feature orientations can improve the split efficiency of standard ensemble methods \citep{masoomi2023explanationsblackboxmodelsbased}. The code can be found in Appendix \ref{app:df_source_code}.

\subsubsection{Algorithm}
We define the forest as an ensemble $\mathcal{E} = \{T_1, T_2, \dots, T_n\}$, where each $T_i$ is a decision tree trained on a modified feature space. The core mechanism is the computation of a directionality vector $\mathbf{d} \in \{-1, 0, 1\}^f$ that rescales features based on their relationship with class means. 

\textbf{Feature Directionality Calculation:} For a training set $(X, y)$, we first compute the mean vector for each class $c \in C$, denoted as $\boldsymbol{\mu}_c$, and the global mean of the dataset $\boldsymbol{\mu}_g$. The directionality vector $\mathbf{d}$ is calculated as the sign of the sum of deviations:
\begin{equation}
    \mathbf{d} = \text{sgn}\left( \sum_{c \in C} (\boldsymbol{\mu}_c - \boldsymbol{\mu}_g) \right)
\end{equation}

\textbf{Directional Transformation:} Before training each tree $T_i$ and during inference, the input features $\mathbf{x}$ are transformed into a directional space $\mathbf{x}_{dir}$ via an element-wise product with the directionality vector:
\begin{equation}
    \mathbf{x}_{dir} = \mathbf{x} \odot \mathbf{d}
\end{equation}
This transformation serves as a primitive form of feature engineering that attempts to orient the features such that splits in the decision trees are more discriminative relative to the global mean.

\begin{figure*}[hb]
    \centering
    \includegraphics[width=1\linewidth]{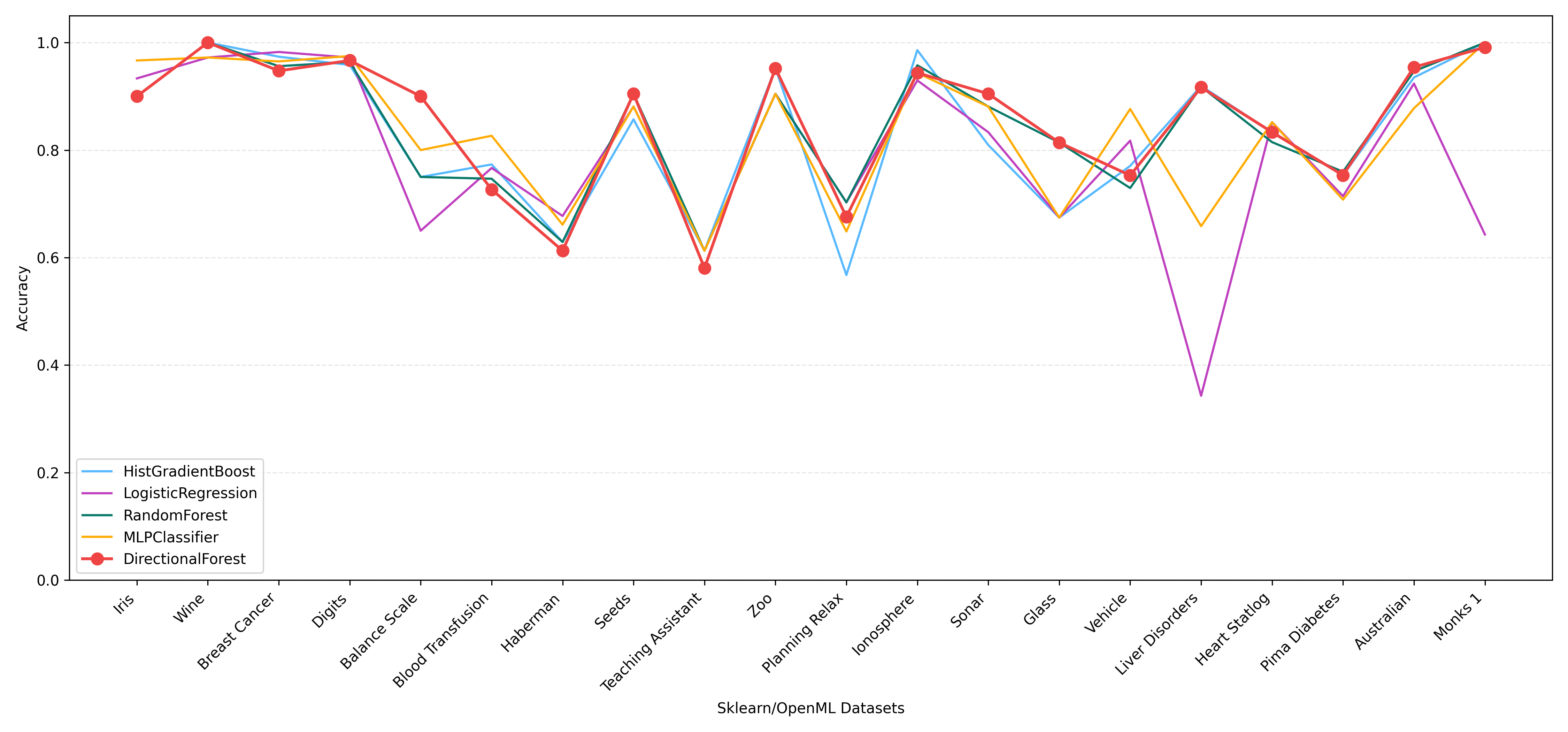}
    \caption{\texttt{DirectionalForest} vs Scikit-Learn Individual Dataset Scores}
    \label{fig:performance_trendline}
\end{figure*}

\textbf{Final Inference:} The forest utilizes a plurality voting mechanism to determine the final class label. For a set of predictions from the ensemble $\{ \hat{y}_1, \hat{y}_2, \dots, \hat{y}_n \}$, the final prediction $\hat{y}$ is:
\begin{equation}
    \hat{y} = \text{mode}(\{ T_i(\mathbf{x} \odot \mathbf{d}) \}_{i=1}^n)
\end{equation}

The algorithm essentially enforces a consistent orientation of the feature space across all learners. By calculating the aggregate sign of class-wise deviations, it attempts to normalize the direction of feature importance, allowing the decision trees to focus on capturing non-linear interactions rather than discovering feature polarity.

\subsubsection{Results}
A closer look at the evaluation metrics shows that the \texttt{DirectionalForest} demonstrates significant stability across high-dimensional datasets. While its linear directionality assumption is simpler than standard Random Forests, it effectively reduces variance in specific manifolds, outperforming Scikit-learn baselines on a majority of the 20 benchmark datasets used for evaluation.

\section{Comparing LLMs within OMEGA}
\label{comparing llms within omega}
Once confirming that this framework could be utilized to generate novel algorithms, we conducted a secondary experiment to determine if the LLM utilized to generate the idea for the algorithms and then subsequently write the code made an impact on how well the model performed.  

In this experiment, we tested four of the most popular code-generation LLMs: \textbf{Anthropic's Claude Sonnet 4.5 model, OpenAI's GPT-4.1 mini, Google's Gemini 2.5 Flash, and xAI's grok-code-fast-1}. (The paper up until this point has used Anthropic's Claude Sonnet 4.5 model exclusively). Table \ref{tab:prompt_benchmarks} shows four of the LLM's performances on the top performing (by average of all LLMs) prompts.

Across the top ten prompts, Gemini performs the best in 6, GPT in 2, and Claude and Grok both in 1. Notably the performance across the four LLMs is quite comparable across all these prompts perhaps alluding to the idea that prompt quality is more important than the actual LLM being used. We do a deeper comparison on whether asking the LLM to iterate on the code vs the prompt performs better in Section \ref{self-improving_prompt_vs_code}

\begin{table*}[h]
\centering
\caption{Scores for Model Generation Prompts Across LLMs \\ (bolded by best performing LLM per prompt)}
\begin{small}
\begin{tabular}{lp{9cm}cccc} 
\toprule
\textbf{PID} & \textbf{Prompt Summary} & \textbf{Gemini} & \textbf{OpenAI} & \textbf{Claude} & \textbf{xAI} \\
\midrule
P01 & Design a framework for incorporating domain knowledge into the Random Forests classifier through the use of abstraction layers and feature engineering. & \textbf{0.9307} & 0.9295 & 0.9202 & 0.9123 \\
P02 & Create a Random Forests classifier that can handle both biased and unbiased data by incorporating a bias-variance decomposition step into the training process. & 0.9236 & 0.9179 & 0.9202 & \textbf{0.9290} \\
P03 & Dynamically adjust the degree of feature sub-sampling randomness for each tree based on the predictability or inherent noise level of its bootstrap sample. & \textbf{0.9281} & 0.9186 & 0.8456 & 0.9202 \\
P04 & Control the degree of randomness (e.g., bootstrap sample size, feature subspace size) for each tree in the ensemble to explicitly target different bias-variance tradeoffs, creating a heterogeneous mix of learners. & 0.9208 & \textbf{0.9268} & 0.8315 & 0.9229 \\
P05 & Design multi-directional split criteria that consider both forward and backward feature dependencies. & \textbf{0.9113} & 0.8849 & 0.8900 & 0.8826 \\
P06 & Create bags with controlled bias-variance profiles by mixing shallow high-bias and deep high-variance trees in learned proportions. & \textbf{0.9177} & 0.8756 & 0.8991 & 0.9141 \\
P07 & Create adaptive random subspace selection where feature subset size varies per tree based on estimated intrinsic dimensionality of the bootstrap sample. & \textbf{0.9322} & 0.8816 & 0.8456 & 0.9134 \\
P08 & Apply entropy-guided feature selection to prioritize features with high information gain. & 0.8901 & 0.8898 & \textbf{0.8914} & 0.8887 \\
P09 & Use hierarchical abstraction layers where shallow trees capture coarse patterns and deep trees capture fine details. & \textbf{0.9161} & 0.9109 & 0.8653 & 0.8578 \\
P10 & Integrate randomness control by varying the degree of noise injected into feature values during tree construction. & 0.8913 & \textbf{0.9036} & 0.8468 & 0.8758 \\
\midrule
& \textbf{Average Performance} & \textbf{0.9162} & \textbf{0.9039} & \textbf{0.8736} & \textbf{0.9017} \\
\bottomrule
\end{tabular}
\end{small}
\label{tab:prompt_benchmarks}
\end{table*}

While Gemini 2.5 Flash was by far the best model at generating executable classification models, this doesn't necessarily correlate with the novelty of code or the ability for these LLM's to ``long-horizon" reason and is solely evaluated on how well their generated classification models perform.

\begin{figure*}[h]
    \centering
    \includegraphics[width=1\linewidth]{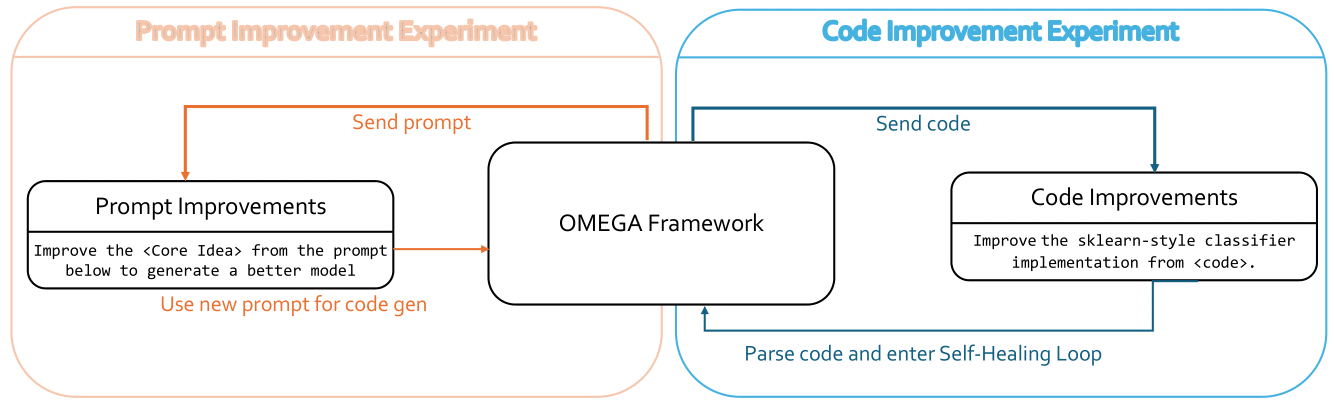}
    \caption{OMEGA Prompt \& Code Improvement Experiments}
    \label{fig:omega_improvements_img}
\end{figure*}

\section{Self-Improving Prompts vs Code}
\label{self-improving_prompt_vs_code}
To further push the capabilities of the OMEGA framework, we implemented a self-improving loop; see Figure \ref{fig:omega_improvements_img}. We iterated on the base architecture, so all the initially generated algorithms would: 
\begin{enumerate}
    \item Generate a new prompt using the initial generation prompt as context 
    \item Generate new code using the old code as context 
\end{enumerate}

In the table below, we present these results of this experiment (conducted independently of the one shown in Figure \ref{comparing llms within omega}). The \textbf{Average Score at Each Step} contains the average score of models generated from the base generation, models generated from prompt improvements, and models generated from code improvements. The \textbf{\% Improvement} contains the percentage improvement from base performance to prompt improvements and base performance to code improvements.

\begin{table}[h]
    \centering
    \label{tab:prompt_benchmarks}
    \caption{Avg Scores and Percentage Improvement (bolded by best strategy for improvement)}
    \begin{tabular}{l ccc cc}
        \toprule
        {\textbf{Model}} & \multicolumn{3}{c}{\textbf{Average Score}} & \multicolumn{2}{c}{\textbf{\% Improvement}} \\
        \cmidrule(lr){2-4} \cmidrule(lr){5-6}
         & \textbf{Base} & \textbf{Prompt} & \textbf{Code} & \textbf{Prompt} & \textbf{Code} \\
        \midrule
        Gemini & 0.885 & 0.894 & 0.890 & \textbf{0.90\%}  & 0.53\%\\
        GPT    & 0.734 & 0.780 & 0.769 & \textbf{4.59\%}  & 3.50\%\\
        Claude & 0.770 & 0.859 & 0.818 & \textbf{8.88\%}  & 4.85\%\\
        Grok   & 0.778 & 0.763 & 0.798 & -1.54\% & \textbf{1.95\%}   \\
        \bottomrule
    \end{tabular}
\end{table}



The results reveal that prompt tuning and optimization outperform pure code optimization for nearly all LLMs, to varying degrees (note that part of the variance in percentage performance will be related to the initial performance between models) \citep{gong2025tuningllmbasedcodeoptimization}. This aligns with the vast amount of prompt-tuning research conducted to prove that fine-tuning prompts can result in significant improvements \citep{shin2023prompt, liu2025beyond, peng2024model}

\section{Conclusion}
In this work, we introduce OMEGA, a framework for end-to-end classification model generation and evaluation. We demonstrate the efficacy of OMEGA by presenting models generated by human and LLM prompting in this framework that are novel and outperform scikit-learn baselines on our evaluation benchmark. We further analyze the best models to use for model generation and the best strategies for self-improvement. We believe that this framework, and similar frameworks that will follow, will open up the door for LLM-generated ideation and code at scale and by technical and non technical engineers.

\newpage



\bibliography{iclr2026_conference}

@book{Hutter2019Automated,
  author    = {Frank Hutter and Lars Kotthoff and Joaquin Vanschoren},
  title     = {Automated Machine Learning: Methods, Systems, Challenges},
  publisher = {Springer},
  year      = {2019}
}

@book{Koza1992Genetic,
  author    = {John R. Koza},
  title     = {Genetic Programming: On the Programming of Computers by Means of Natural Selection},
  publisher = {MIT Press},
  year      = {1992}
}

@article{Zoph2016Neural,
  author  = {Barret Zoph and Quoc V. Le},
  title   = {Neural Architecture Search with Reinforcement Learning},
  journal = {arXiv preprint arXiv:1611.01578},
  year    = {2016}
}

@inproceedings{Snoek2012Practical,
  author    = {J. Snoek and H. Larochelle and R. P. Adams},
  title     = {Practical Bayesian optimization of machine learning algorithms},
  booktitle = {Advances in Neural Information Processing Systems},
  pages     = {2951--2959},
  year      = {2012}
}

@inproceedings{Finn2017Model,
  author    = {Chelsea Finn and Pieter Abbeel and Sergey Levine},
  title     = {Model-agnostic meta-learning for fast adaptation of deep networks},
  booktitle = {Proceedings of the 34th International Conference on Machine Learning},
  volume    = {70},
  pages     = {1126--1135},
  publisher = {PMLR},
  year      = {2017}
}

@article{elsken2019neuralarchserach,
  author  = {Thomas Elsken and Jan Hendrik Metzen and Frank Hutter},
  title   = {Neural Architecture Search: A Survey},
  journal = {Journal of Machine Learning Research},
  volume  = {20},
  number  = {55},
  pages   = {1--21},
  year    = {2019}
}

@inproceedings{Real2020AlphaEvolve,
  author    = {Esteban Real and Chen Liang and David So and Quoc Le},
  title     = {AutoML-Zero: Evolving Machine Learning Algorithms From Scratch},
  booktitle = {Proceedings of the 37th International Conference on Machine Learning},
  series    = {PMLR},
  volume    = {119},
  pages     = {8007--8017},
  year      = {2020}
}

@article{Fawzi2022AlphaTensor,
  author  = {Alhussein Fawzi and others},
  title   = {Discovering faster matrix multiplication algorithms with reinforcement learning},
  journal = {Nature},
  volume  = {610},
  number  = {7930},
  pages   = {47--53},
  year    = {2022}
}

@article{Romera2024FunSearch,
  author  = {Bernardino Romera-Paredes and others},
  title   = {Mathematical discoveries from program search with large language models},
  journal = {Nature},
  volume  = {625},
  number  = {7995},
  pages   = {468--475},
  year    = {2024}
}

@article{Lu2024AIScientist,
  author  = {Chris Lu and others},
  title   = {The AI Scientist: Towards Fully Automated Machine Learning Scientific Discovery},
  journal = {arXiv preprint arXiv:2408.06292},
  year    = {2024}
}

@article{Akiba2024Sakana,
  author  = {Takuya Akiba and Makoto Shing and Yasuhiro Majima and Shinya Kaneko},
  title   = {Evolutionary Optimization of Model Merging Recipes},
  journal = {arXiv preprint arXiv:2403.13187},
  year    = {2024}
}

@article{chen2021evaluating,
  author  = {Mark Chen and others},
  title   = {Evaluating Large Language Models Trained on Code},
  journal = {arXiv preprint arXiv:2107.03374},
  year    = {2021}
}

@misc{openai2023code,
  author       = {OpenAI},
  title        = {ChatGPT Code Interpreter},
  howpublished = {OpenAI Blog},
  year         = {2023}
}

@misc{openai2023assistants,
  author       = {OpenAI},
  title        = {Assistants API},
  howpublished = {OpenAI Documentation},
  year         = {2023}
}

@article{Vilalta2002Perspective,
  author  = {Ricardo Vilalta and Youssef Drissi},
  title   = {A Perspective View and Survey of Meta-Learning},
  journal = {Artificial Intelligence Review},
  volume  = {18},
  number  = {2},
  pages   = {77--95},
  year    = {2002}
}

@inproceedings{Colton2008Creativity,
  author    = {Simon Colton and Geraint A. Wiggins},
  title     = {Computational creativity: The final frontier?},
  booktitle = {Proceedings of the 20th European Conference on Artificial Intelligence},
  pages     = {21--26},
  year      = {2008}
}

@article{Pedregosa2011Scikit,
  author  = {Fabian Pedregosa and others},
  title   = {Scikit-learn: Machine Learning in Python},
  journal = {Journal of Machine Learning Research},
  volume  = {12},
  pages   = {2825--2830},
  year    = {2011}
}

@inproceedings{Buitinck2013API,
  author    = {Lars Buitinck and Gilles Louppe and Mathieu Blondel and Fabian Pedregosa and Andreas Mueller and Olivier Grisel and Vlad Niculae and Peter Prettenhofer and Alexandre Gramfort and Jaques Grobler and Robert Layton and Jake VanderPlas and Arnaud Joly and Brian Holt and Ga{\"e}l Varoquaux},
  title     = {API design for machine learning software: experiences from the scikit-learn project},
  booktitle = {ECML PKDD Workshop: Languages for Data Mining and Machine Learning},
  pages     = {108--122},
  year      = {2013}
}

@inproceedings{Gundersen2018Reproducible,
  author    = {Odd Erik Gundersen and Sigbjørn Kjensmo},
  title     = {State of the art: Reproducibility in artificial intelligence},
  booktitle = {Proceedings of the Thirty-Second AAAI Conference on Artificial Intelligence},
  pages     = {1644--1651},
  year      = {2018}
}

@book{Aho2006Compilers,
  author    = {Alfred V. Aho and Monica S. Lam and Ravi Sethi and Jeffrey D. Ullman},
  title     = {Compilers: Principles, Techniques, and Tools},
  publisher = {Pearson/Addison Wesley},
  edition   = {2nd},
  year      = {2006}
}

@article{Vanschoren2014OpenML,
  author  = {Joaquin Vanschoren and Jan N. van Rijn and Bernd Bischl and Luis Torgo},
  title   = {OpenML: Networked science in machine learning},
  journal = {SIGKDD Explorations},
  volume  = {15},
  number  = {2},
  pages   = {49--60},
  year    = {2014}
}

@article{Demvsar2006Statistical,
  author  = {Janez Dem\v{s}ar},
  title   = {Statistical Comparisons of Classifiers over Multiple Data Sets},
  journal = {Journal of Machine Learning Research},
  volume  = {7},
  pages   = {1--30},
  year    = {2006}
}

@article{Wolpert1992Stacked,
  author  = {David H. Wolpert},
  title   = {Stacked Generalization},
  journal = {Neural Networks},
  volume  = {5},
  number  = {2},
  pages   = {241--259},
  year    = {1992}
}

@article{shin2023prompt,
  author  = {Jiho Shin and Clark Tang and Tahmineh Mohati and Maleknaz Nayebi and Song Wang and Hadi Hemmati},
  title   = {Prompt Engineering or Fine-Tuning: An Empirical Assessment of LLMs for Code},
  journal = {arXiv preprint arXiv:2310.10508},
  year    = {2023}
}

@article{liu2025beyond,
  author  = {Yuanye Liu and Jiahang Xu and Li Lyna Zhang and Qi Chen and Xuan Feng and Yang Chen and Zhongxin Guo and Yuqing Yang and Cheng Peng},
  title   = {Beyond Prompt Content: Enhancing LLM Performance via Content-Format Integrated Prompt Optimization},
  journal = {arXiv preprint arXiv:2502.04295},
  year    = {2025}
}

@article{peng2024model,
  author  = {Cheng Peng and Xi Yang and Kaleb E. Smith and Zehao Yu and Aokun Chen and Jiang Bian and Yonghui Wu},
  title   = {Model Tuning or Prompt Tuning? A Study of Large Language Models for Clinical Concept and Relation Extraction},
  journal = {Journal of Biomedical Informatics},
  volume  = {153},
  pages   = {104630},
  year    = {2024}
}

@inproceedings{Wang_2025_unexamined_code,
   title={RefleXGen:The unexamined code is not worth using},
   url={http://dx.doi.org/10.1109/ICASSP49660.2025.10890824},
   DOI={10.1109/icassp49660.2025.10890824},
   booktitle={ICASSP 2025 - 2025 IEEE International Conference on Acoustics, Speech and Signal Processing (ICASSP)},
   publisher={IEEE},
   author={Wang, Bin and Li, Hui and Liu, AoFan and Yang, BoTao and Yang, Ao and Zhong, YiLu and Huang, Weixiang and Huang, Runhuai and Zeng, Weimin and Zhang, Yanping},
   year={2025},
   month=apr, pages={1–5} }

@misc{gong2025tuningllmbasedcodeoptimization,
      title={Tuning LLM-based Code Optimization via Meta-Prompting: An Industrial Perspective}, 
      author={Jingzhi Gong and Rafail Giavrimis and Paul Brookes and Vardan Voskanyan and Fan Wu and Mari Ashiga and Matthew Truscott and Mike Basios and Leslie Kanthan and Jie Xu and Zheng Wang},
      year={2025},
      eprint={2508.01443},
      archivePrefix={arXiv},
      primaryClass={cs.SE},
      url={https://arxiv.org/abs/2508.01443}, 
}

@book{brazdil2009metalearning,
  title={Metalearning: Applications to Data Mining},
  author={Brazdil, Pavel and Giraud-Carrier, Christophe and Soares, Carlos and Vilalta, Ricardo},
  year={2009},
  publisher={Springer},
  address={Berlin, Heidelberg},
  series={Cognitive Technologies},
  isbn={978-3-540-73262-4},
  doi={10.1007/978-3-540-73263-1}
}

@misc{masoomi2023explanationsblackboxmodelsbased,
      title={Explanations of Black-Box Models based on Directional Feature Interactions}, 
      author={Aria Masoomi and Davin Hill and Zhonghui Xu and Craig P Hersh and Edwin K. Silverman and Peter J. Castaldi and Stratis Ioannidis and Jennifer Dy},
      year={2023},
      eprint={2304.07670},
      archivePrefix={arXiv},
      primaryClass={cs.LG},
      url={https://arxiv.org/abs/2304.07670}, 
}
\bibliographystyle{iclr2026_conference}

\appendix
\section{Appendix}
\subsection{Source Code: \texttt{MetaSynthesisClassifier}}
\label{app:metasynthesis_source_code}
\begin{lstlisting}[language=Python]
import numpy as np
from sklearn.base import BaseEstimator, ClassifierMixin, clone
from sklearn.utils.validation import check_X_y, check_array, check_is_fitted
from sklearn.utils.multiclass import unique_labels
from sklearn.model_selection import cross_val_predict
from sklearn.linear_model import LogisticRegression
from sklearn.ensemble import RandomForestClassifier
from sklearn.tree import DecisionTreeClassifier

class MetaSynthesisClassifier(BaseEstimator, ClassifierMixin):
    def __init__(self, base_estimators=None, meta_estimator=None, cv=5, 
                 use_probas=True, use_original_features=False):
        self.base_estimators = base_estimators
        self.meta_estimator = meta_estimator
        self.cv = cv
        self.use_probas = use_probas
        self.use_original_features = use_original_features
    
    def fit(self, X, y):
        X, y = check_X_y(X, y)
        self.classes_ = unique_labels(y)
        self.n_features_in_ = X.shape[1]
        
        if self.base_estimators is None:
            self.base_estimators_ = [
                LogisticRegression(max_iter=1000, random_state=42),
                RandomForestClassifier(n_estimators=100, random_state=42),
                DecisionTreeClassifier(random_state=42)
            ]
        else:
            self.base_estimators_ = [clone(est) for est in self.base_estimators]
        
        if self.meta_estimator is None:
            self.meta_estimator_ = LogisticRegression(max_iter=1000, random_state=42)
        else:
            self.meta_estimator_ = clone(self.meta_estimator)
        
        meta_features = self._generate_meta_features(X, y)
        
        for estimator in self.base_estimators_:
            estimator.fit(X, y)
        
        if self.use_original_features:
            meta_features = np.hstack([X, meta_features])
        
        self.meta_estimator_.fit(meta_features, y)
        return self
    
    def predict(self, X):
        check_is_fitted(self)
        X = check_array(X)
        meta_features = self._generate_meta_features_predict(X)
        
        if self.use_original_features:
            meta_features = np.hstack([X, meta_features])
        
        return self.meta_estimator_.predict(meta_features)
    
    def predict_proba(self, X):
        check_is_fitted(self)
        X = check_array(X)
        meta_features = self._generate_meta_features_predict(X)
        
        if self.use_original_features:
            meta_features = np.hstack([X, meta_features])
        
        if hasattr(self.meta_estimator_, 'predict_proba'):
            return self.meta_estimator_.predict_proba(meta_features)
        else:
            raise AttributeError("Meta estimator does not support predict_proba")
    
    def _generate_meta_features(self, X, y):
        meta_features_list = []
        for estimator in self.base_estimators_:
            if self.use_probas and hasattr(estimator, 'predict_proba'):
                cv_preds = cross_val_predict(
                    estimator, X, y, cv=self.cv, method='predict_proba'
                )
                meta_features_list.append(cv_preds)
            else:
                cv_preds = cross_val_predict(estimator, X, y, cv=self.cv)
                meta_features_list.append(cv_preds.reshape(-1, 1))
        
        return np.hstack(meta_features_list)
    
    def _generate_meta_features_predict(self, X):
        meta_features_list = []
        for estimator in self.base_estimators_:
            if self.use_probas and hasattr(estimator, 'predict_proba'):
                preds = estimator.predict_proba(X)
                meta_features_list.append(preds)
            else:
                preds = estimator.predict(X)
                meta_features_list.append(preds.reshape(-1, 1))
        
        return np.hstack(meta_features_list)
\end{lstlisting}

\newpage
\subsection{Implementation of \texttt{DirectionalForest}}
\label{app:df_source_code}
\begin{lstlisting}[language=Python]
    import numpy as np
from sklearn.base import BaseEstimator, ClassifierMixin
from sklearn.utils.validation import check_X_y, check_array, check_is_fitted
from sklearn.utils.multiclass import unique_labels
from sklearn.tree import DecisionTreeClassifier

class DirectionalForest(BaseEstimator, ClassifierMixin):
    def __init__(self, n_estimators=100, max_depth=None, min_samples_split=2, 
                 min_samples_leaf=1, max_features='sqrt', random_state=None):
        self.n_estimators = n_estimators
        self.max_depth = max_depth
        self.min_samples_split = min_samples_split
        self.min_samples_leaf = min_samples_leaf
        self.max_features = max_features
        self.random_state = random_state
        
    def fit(self, X, y):
        X, y = check_X_y(X, y)
        
        self.classes_ = unique_labels(y)
        
        self.feature_directions_ = self._calculate_feature_directions(X, y)
        
        self.estimators_ = []
        
        for _ in range(self.n_estimators):
            tree = self._grow_tree(X, y)
            self.estimators_.append(tree)
        
        return self
    
    def predict(self, X):
        check_is_fitted(self)
        
        X = check_array(X)
        
        X_directional = X * self.feature_directions_
        
        predictions = np.array([tree.predict(X_directional) for tree in self.estimators_])
        
        return np.apply_along_axis(lambda x: np.argmax(np.bincount(x)), axis=0, arr=predictions)
    
    def _grow_tree(self, X, y):
        tree = DecisionTreeClassifier(
            max_depth=self.max_depth,
            min_samples_split=self.min_samples_split,
            min_samples_leaf=self.min_samples_leaf,
            max_features=self.max_features,
            random_state=self.random_state
        )
        
        X_directional = X * self.feature_directions_
        
        tree.fit(X_directional, y)
        
        return tree
    
    def _calculate_feature_directions(self, X, y):
        class_means = [np.mean(X[y == c], axis=0) for c in self.classes_]
        
        overall_mean = np.mean(X, axis=0)
        
        directions = np.sign(np.sum([cm - overall_mean for cm in class_means], axis=0))
        
        return directions
\end{lstlisting}

\end{document}